\documentclass[cameraready]{Interspeech}
\usepackage{tikz}
\usetikzlibrary{arrows.meta, positioning, calc, shadows, shapes.geometric}

\usepackage{pifont}
\usepackage{tcolorbox}
\title{DLLM-TTS: Block Discrete Diffusion Language Model for Text-to-Speech Synthesis}

\author[affiliation={1}, correspondingauthor]{Wasim}{Madha}
\author[affiliation={1}]{Nityanand}{Mathur}
\author[affiliation={1}]{Hamees}{Sayed}
\author[affiliation={1}]{Apoorv}{Singh}
\author[affiliation={1}]{Sameer}{Khurana}
\author[affiliation={1}]{Akshat}{Mandloi}
\author[affiliation={1}]{Sudarshan}{Kamath}

\address{
    $^1$ Smallest.ai
}

\email{wasim@smallest.ai, nityanand.mathur@iiitg.ac.in, hamees@smallest.ai, apoorv.singh@smallest.ai, sameerkhurana10@gmail.com, akshat@smallest.ai, sudarshan@smallest.ai}

\keywords{discrete diffusion, text-to-speech, block diffusion, neural codec, language model}

\usepackage{comment}
\usepackage{tikz}
\usetikzlibrary{arrows.meta}
\definecolor{codecteal}{RGB}{0,128,128}
\definecolor{seqviolet}{RGB}{120,80,160}
\definecolor{mainblue}{RGB}{41, 128, 185}
\definecolor{accentteal}{RGB}{22, 160, 133}
\definecolor{accentorange}{RGB}{230, 126, 34}
\definecolor{accentred}{RGB}{192, 57, 43}
\definecolor{accentviolet}{RGB}{142, 68, 173}
\newcommand{\myparagraph}[1]{\vspace{0pt}\noindent{\bf #1}}

\makeatletter
\newcommand*\patchAmsMathEnvironmentForLineno[1]{%
  \expandafter\let\csname old#1\expandafter\endcsname\csname #1\endcsname
  \expandafter\let\csname oldend#1\expandafter\endcsname\csname end#1\endcsname
  \renewenvironment{#1}%
   {\linenomath\csname old#1\endcsname}%
   {\csname oldend#1\endcsname\endlinenomath}}%
\newcommand*\patchBothAmsMathEnvironmentsForLineno[1]{%
  \patchAmsMathEnvironmentForLineno{#1}%
  \patchAmsMathEnvironmentForLineno{#1*}}%
\AtBeginDocument{%
\patchBothAmsMathEnvironmentsForLineno{equation}%
\patchBothAmsMathEnvironmentsForLineno{align}%
\patchBothAmsMathEnvironmentsForLineno{gather}%
\patchBothAmsMathEnvironmentsForLineno{multline}%
}
\makeatother

\begin{document}

\maketitle

\begin{abstract}
Current text-to-speech systems face a trade-off: autoregressive codec language models produce highly intelligible speech but require large-scale models and training data and decode tokens sequentially, while non-autoregressive approaches improve speed at the cost of linguistic accuracy. We present DLLM-TTS, a framework that formulates TTS as conditional block discrete diffusion over X-Codec2 neural audio codec tokens. The model decomposes sequences into blocks and applies masked diffusion within each block while processing blocks sequentially, learning both local acoustic coherence and global text-speech alignment. During inference, parallel token prediction within blocks enables efficient generation with a real-time factor (RTF) of 0.15. A 0.6B-parameter model trained on 20K hours achieves competitive performance on the Seed-TTS-eval benchmark, demonstrating that block discrete diffusion language models enable practical and data-efficient speech synthesis with parallel generation.
\end{abstract}

\section{Introduction}

Text-to-speech synthesis faces a fundamental efficiency-quality trade-off. Autoregressive codec language models~\cite{wang2023valle,chen2024valle2,ye2025llasa} achieve high-quality zero-shot synthesis but require 60K--250K hours of data and generate tokens sequentially, incurring high latency. Non-autoregressive approaches based on flow matching~\cite{le2024voicebox,chen2024f5tts} and diffusion~\cite{ju2024naturalspeech3,li2023styletts2} enable parallel generation but typically need explicit duration modeling or struggle with text-speech alignment, leading to word skipping or repetition errors.

Recent masked discrete diffusion language models~\cite{sahoo2024mdlm,nie2025llada} have shown that discrete diffusion can match autoregressive quality for text generation with parallel decoding. BD3-LM~\cite{arriola2025bd3lm} further introduces block decomposition, interpolating between autoregressive and fully parallel generation with flexible speed-quality trade-offs. However, these methods have only been explored for text; their application to conditional speech generation over discrete codec tokens remains unexplored.

We present \textbf{DLLM-TTS}, which adapts block discrete diffusion to conditional speech generation. Speech exhibits strong local acoustic coherence at the phoneme level while requiring longer-range dependencies for text-speech alignment, a structure naturally suited to block diffusion. We model speech as block discrete diffusion over X-Codec2~\cite{ye2024xcodec} codec tokens: sequences are decomposed into blocks, and the model reconstructs masked tokens within each block conditioned on text, enabling parallel prediction within blocks while preserving sequential dependencies across them. The masked diffusion objective additionally provides implicit data augmentation, as each sequence is observed under diverse masking patterns, improving data efficiency over autoregressive training.

\myparagraph{Contributions.}
\ding{182}~We introduce block discrete diffusion for TTS, modeling speech synthesis as conditional masked token reconstruction over codec tokens with block decomposition, the first application of block discrete diffusion to conditional speech generation.
\ding{183}~We employ staircase attention to jointly capture local acoustic coherence within blocks and global text-speech alignment across blocks, without explicit duration modeling.
\ding{184}~We achieve competitive intelligibility on the Seed-TTS benchmark~\cite{anastassiou2024seedtts} using only 20K hours of data, a 3--12$\times$ reduction over autoregressive systems.
\ding{185}~A 0.6B-parameter model achieves an RTF of 0.15 with strong zero-shot speaker similarity.
\section{Related Work}

\myparagraph{Autoregressive Codec Language Models.}
A dominant paradigm in TTS formulates speech synthesis as language modeling over discrete neural codec tokens. VALL-E~\cite{wang2023valle} pioneered this approach by training on 60K hours of speech, achieving strong zero-shot synthesis from a 3-second prompt. VALL-E~2~\cite{chen2024valle2} introduced repetition-aware sampling and grouped code modeling to reach human parity on standard benchmarks. LLASA~\cite{ye2025llasa} demonstrated that scaling LLaMA-based architectures with X-Codec2~\cite{ye2024xcodec} up to 8B parameters yields consistent quality gains. DiTAR~\cite{jia2025ditar} combines a language model with a diffusion transformer in a patch-based autoregressive framework for continuous-valued speech generation. While these systems achieve high quality, sequential token generation introduces latency that limits real-time applications.

\myparagraph{Non-Autoregressive Speech Synthesis.}
To overcome the latency bottleneck, non-autoregressive methods generate speech in parallel. Voicebox~\cite{le2024voicebox} applies flow matching over mel-spectrograms conditioned on text and surrounding audio context. F5-TTS~\cite{chen2024f5tts} simplifies the pipeline by using flow matching with a Diffusion Transformer (DiT), eliminating duration models and phoneme alignment. NaturalSpeech~3~\cite{ju2024naturalspeech3} employs factorized diffusion over disentangled speech attributes (content, prosody, timbre). StyleTTS~2~\cite{li2023styletts2} achieves human-level quality through style diffusion and adversarial training with speech language model discriminators. CosyVoice~\cite{du2024cosyvoice} combines an LLM-based text-to-token stage with conditional flow matching for token-to-speech synthesis. SoundStorm~\cite{borsos2023soundstorm} uses confidence-based parallel decoding over RVQ tokens for fast generation but struggles with text-speech alignment when trained from scratch. MegaTTS~3~\cite{jiang2025megatts3} proposes sparse alignment with a latent diffusion transformer for robust zero-shot synthesis. These approaches typically require explicit duration prediction, frame-aligned annotations, or multi-stage pipelines, which complicate training data preparation.

\myparagraph{Discrete Diffusion Language Models.}
Masked discrete diffusion has recently emerged as a viable alternative to autoregressive modeling for text generation. MDLM~\cite{sahoo2024mdlm} derives a simplified objective based on masked language modeling losses, achieving strong performance among diffusion-based language models. LLaDA~\cite{nie2025llada} scales masked diffusion to 8B parameters, demonstrating competitive performance with autoregressive LLMs on in-context learning and instruction following. BD3-LM~\cite{arriola2025bd3lm} extends MDLM with block decomposition, interpolating between fully parallel diffusion and autoregressive generation while enabling variable-length output and KV caching. However, these methods have primarily been studied for text generation. Their application to conditional speech generation, where both acoustic coherence and text–speech alignment must be modeled jointly over discrete codec tokens, remains largely unexplored. Our work addresses this gap by adapting block discrete diffusion for text-to-speech synthesis.

\section{Methodology}

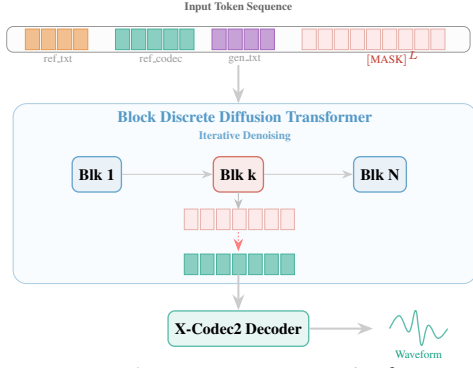
\begin{figure}[t]
\centering
\scalebox{0.85}{\definecolor{mainblue}{RGB}{41, 128, 185}
\definecolor{softblue}{RGB}{235, 245, 251}
\definecolor{accentteal}{RGB}{22, 160, 133}
\definecolor{softteal}{RGB}{232, 248, 245}
\definecolor{softorange}{RGB}{254, 245, 231}
\definecolor{accentorange}{RGB}{230, 126, 34}
\definecolor{softred}{RGB}{253, 235, 236}
\definecolor{accentred}{RGB}{192, 57, 43}
\definecolor{softviolet}{RGB}{245, 238, 248}
\definecolor{accentviolet}{RGB}{142, 68, 173}

\begin{tikzpicture}[
>=Stealth,
node distance=0.8cm,
mod/.style={rectangle, rounded corners=3pt, draw=#1!70, fill=#1!10,
minimum height=0.55cm, text centered, font=\scriptsize\bfseries,
inner sep=3pt, line width=0.6pt},
tok/.style={rectangle, minimum width=0.22cm, minimum height=0.32cm,
fill=#1!50, draw=#1!80, line width=0.2pt},
mtok/.style={rectangle, minimum width=0.22cm, minimum height=0.32cm,
fill=softred, draw=accentred!40, line width=0.3pt},
lbl/.style={font=\tiny\bfseries, text=black!60},
sublbl/.style={font=\fontsize{5}{6}\selectfont, text=black!40}
]

\node[lbl] at (3.8,5.6) {Input Token Sequence};

\draw[rounded corners=3pt, fill=gray!2, draw=black!40]
(0.2,4.9) rectangle (7.4,5.3);

\foreach \i in {0,...,3}
\node[tok=accentorange] at (0.6+\i*0.25,5.1) {};

\node[sublbl] at (1.0,4.8) {ref\_txt};

\foreach \i in {0,...,4}
\node[tok=accentteal] at (2.0+\i*0.25,5.1) {};

\node[sublbl] at (2.6,4.8) {ref\_codec};

\foreach \i in {0,...,3}
\node[tok=accentviolet] at (3.5+\i*0.25,5.1) {};

\node[sublbl] at (3.9,4.8) {gen\_txt};

\foreach \i in {0,...,8}
\node[mtok] at (4.9+\i*0.25,5.1) {};

\node[sublbl,text=accentred] at (6.2,4.8) {$[\text{MASK}]^L$};

\draw[->,line width=0.9pt,color=gray!40]
(3.8,4.7) -- (3.8,4.1);

\draw[rounded corners=6pt, fill=softblue!30, draw=mainblue!40]
(0.3,1.3) rectangle (7.5,4.1);

\node[font=\scriptsize\bfseries,text=mainblue!80]
at (3.9,3.9)
{Block Discrete Diffusion Transformer};

\node[lbl,text=mainblue!70]
at (3.9,3.6)
{Iterative Denoising};

\node[mod=mainblue] (b1) at (1.6,3.0) {Blk 1};
\node[mod=accentred] (bk) at (3.8,3.0) {Blk k};
\node[mod=mainblue] (bn) at (6.0,3.0) {Blk N};

\draw[->,gray!40] (b1) -- (bk);
\draw[->,gray!40] (bk) -- (bn);

\draw[->,gray!50] (3.8,2.725) -- (3.8,2.45);

\foreach \i in {0,...,6}
\node[mtok,minimum width=0.18cm]
at (3.065+\i*0.25,2.3) {};

\draw[->,red!50,densely dotted,line width=0.6pt] (3.8,2.1) -- (3.8,1.8);

\foreach \i in {0,...,6}
\node[tok=accentteal,minimum width=0.18cm]
at (3.065+\i*0.25,1.6) {};

\node[mod=accentteal] (dec) at (3.8,0.6) {X-Codec2 Decoder};
\node[minimum width=1.4cm, minimum height=1.0cm] (sp) at (6.6,0.6) {};
\draw[accentteal!70, semithick] plot[smooth, tension=0.7] coordinates {
    ([xshift=-0.45cm]sp.center) ([xshift=-0.3cm, yshift=0.2cm]sp.center)
    ([xshift=-0.15cm, yshift=-0.05cm]sp.center) ([xshift=-0.05cm, yshift=0.28cm]sp.center)
    (sp.center) ([xshift=0.05cm, yshift=-0.26cm]sp.center)
    ([xshift=0.15cm, yshift=0.05cm]sp.center) ([xshift=0.3cm, yshift=-0.15cm]sp.center)
    ([xshift=0.45cm]sp.center)
};
\node[sublbl, text=accentteal] at ([yshift=-0.4cm]sp.center) {Waveform};

\draw[->,line width=0.9pt,gray!40]
(3.8,1.445) -- (dec);

\draw[->,line width=0.9pt,gray!40]
(dec) -- (sp);

\end{tikzpicture}}
\caption{Overview of DLLM-TTS. \textbf{Text} and \textbf{reference codec tokens} are concatenated with \textbf{generation text} and \textbf{\texttt{[MASK]}$^L$} targets, then processed by the \textbf{Block Discrete Diffusion Transformer} using staircase attention (bidirectional/causal). Each block is \textbf{iteratively denoised} and decoded by \textbf{X-Codec2} into speech.}
\label{fig:architecture}
\end{figure}

\subsection{Background}

\myparagraph{Neural Audio Codecs.}
Neural audio codecs compress continuous audio waveforms into discrete token sequences through learned quantization. Traditional residual vector quantization (RVQ) approaches like EnCodec~\cite{defossez2022encodec} employ multiple codebook layers to progressively refine representations, producing parallel token streams. In contrast, X-Codec2~\cite{ye2024xcodec} adopts a unified semantic-acoustic architecture: a semantic encoder (Wav2Vec2-BERT) captures linguistic content while an acoustic encoder preserves fine-grained audio characteristics. These representations are fused and quantized using single-stage Finite Scalar Quantization (FSQ) with vocabulary size $|\mathcal{V}| = 6561$, producing a single token stream at frame rate $f_r = 50$\,Hz. This yields $L = f_r \cdot d_{\text{sec}}$ codec tokens for $d_{\text{sec}}$ seconds of audio, simplifying integration with transformer language models by avoiding multiple parallel codebook streams.

\myparagraph{Masked Discrete Diffusion.}
Masked diffusion~\cite{sahoo2024mdlm} defines a forward process that progressively corrupts discrete sequences by replacing tokens with a special \texttt{[MASK]} token, and a reverse process that learns to denoise the corrupted observations. For a sequence $\mathbf{x} = (x_1, \ldots, x_L)$ where each $x_i \in \mathcal{V}$, the forward process at continuous timestep $t \in [0,1]$ corrupts each token independently:
\begin{equation}
q(\mathbf{z}_t \mid \mathbf{x}) = \prod_{i=1}^{L} \big[\alpha_t \cdot \delta(z_t^i = x_i) + (1 - \alpha_t) \cdot \delta(z_t^i = \texttt{[MASK]})\big]
\label{eq:forward}
\end{equation}
where $\alpha_t \in [0,1]$ is a monotonically decreasing schedule function. At $t=0$ the sequence is fully clean ($\alpha_0=1$), while at $t=1$ it is fully masked ($\alpha_1 \approx 0$). The model learns the reverse process $p_\theta(\mathbf{x} \mid \mathbf{z}_t, \mathbf{c})$ to predict original tokens given corrupted observations $\mathbf{z}_t$ and conditioning $\mathbf{c}$.

\myparagraph{Block Discrete Diffusion.}
BD3-LM~\cite{arriola2025bd3lm} extends masked diffusion by decomposing sequences into blocks of size $B$. Rather than processing the entire sequence uniformly, block diffusion generates blocks sequentially while allowing parallel prediction within each block. During training, the model receives concatenated input $[\mathbf{x}_t \oplus \mathbf{x}_0]$ where $\mathbf{x}_t$ contains masked tokens and $\mathbf{x}_0$ contains clean tokens from previous blocks. A specialized \textit{staircase attention} mask enforces: (1)~bidirectional attention within each block for local coherence, (2)~causal attention from noised blocks to previous clean blocks for sequential context, and (3)~full causal attention within clean blocks. This architecture interpolates between fully parallel diffusion ($B = L$) and autoregressive generation ($B = 1$), enabling a flexible trade-off between generation speed and sequential dependency modeling.

\subsection{Block Discrete Diffusion for Speech}

We formulate text-to-speech synthesis as conditional block discrete diffusion over codec token sequences. Given a text input and speaker prompt, the model generates speech by iteratively denoising masked codec tokens through a block-based generation process.

\myparagraph{Sequence Decomposition.}
A codec token sequence $\mathbf{x} = (x_1, \ldots, x_L)$ is partitioned into $K$ contiguous blocks of size $B$:
\begin{equation}
\mathbf{x} = [\mathbf{x}^{(1)}, \ldots, \mathbf{x}^{(K)}], \quad K = \lceil L / B \rceil, \quad \mathbf{x}^{(k)} \in \mathcal{V}^{B}
\label{eq:blocks}
\end{equation}
where $B = 32$ tokens ($\sim$0.64\,s of audio at $f_r = 50$\,Hz).

\myparagraph{Forward Process.}
For a given diffusion timestep $t \in [0,1]$, tokens within the target block are independently masked with probability $1 - \alpha_t$ where $\alpha_t = 1 - t$ is a linear schedule. Concretely, for each position $i$ in block $k$, a binary mask is sampled as $m_i^{(k)} \sim \text{Bernoulli}(1 - \alpha_t)$, and the corrupted token is:
\begin{equation}
z_t^i = (1 - m_i^{(k)})\, x_i + m_i^{(k)}\, \texttt{[MASK]}
\label{eq:masking}
\end{equation}
i.e., positions where $m_i^{(k)} = 1$ are replaced with \texttt{[MASK]} while others remain clean. The resulting masked sequence $\mathbf{z}_t$ forms the corrupted observation that the model learns to reconstruct.

\myparagraph{Reverse Process and Training.}
The model predicts original unmasked tokens given $\mathbf{z}_t$, conditioned on both text and speaker information. We adopt staircase attention, where position $i$ attends to position $j$ according to the binary mask $\mathbf{A}_{ij} \in \{0,1\}$, defined as:
\begin{equation}
\mathbf{A}_{ij} = \mathbf{1}[\,C_1 \vee C_2 \vee C_3\,]
\label{eq:attn}
\end{equation}
with three conditions: $C_1{:}$ $\; b_i {=} b_j \wedge b_i {\leq} b^*$ (bidirectional within noised blocks), $C_2{:}\; b_j {<} b_i$ (causal across blocks), and $C_3{:}\; b_i {=} b_j \wedge b_i {>} b^* \wedge j {\leq} i$ (causal within clean blocks), where $b_i = \lceil i/B \rceil$ maps position $i$ to its block index, $b^*$ is the current noised block, and $\mathbf{1}[\cdot]$ is the indicator function. This enables bidirectional attention within each noised block for local coherence, causal attention to previous clean blocks for sequential context, and causal attention within clean blocks for autoregressive conditioning. The training objective minimizes cross-entropy loss over masked positions:
\begin{equation}
\mathcal{L} = \mathbb{E}_{t \sim \mathcal{U}[0,1],\, \mathbf{x},\, \mathbf{z}_t \sim q(\cdot|\mathbf{x})} \bigg[ -\sum_{i \in \mathcal{M}_t} \log p_\theta(x_i \mid \mathbf{z}_t, \mathbf{c}) \bigg]
\label{eq:loss}
\end{equation}
where $\mathcal{M}_t$ denotes the set of masked positions at timestep $t$, and $\mathbf{c}$ represents conditioning information (text and speaker prompt).

\myparagraph{Variable-Length Handling.}
To handle variable-length sequences, we introduce an end-of-sequence (EOS) token. All tokens following EOS are set to EOS, which simplifies the masking process and eliminates the need for explicit length prediction, as the model learns to generate EOS when synthesis is complete.

\subsection{Model Architecture}

Our model is a 0.6B-parameter transformer initialized from Qwen2~\cite{yang2024qwen2} and adapted for block discrete diffusion. Table~\ref{tab:arch} summarizes the architecture. The model uses Rotary Position Embeddings (RoPE)~\cite{su2024rope} for efficient long-context modeling.

\begin{table}[t]
\centering
\caption{Model architecture hyperparameters.}
\label{tab:arch}
\renewcommand{\arraystretch}{1.1}
\begin{tabular}{lc}
\toprule
Hyperparameter & Value \\
\midrule
Layers & 28 \\
Hidden dim ($d$) & 896 \\
Attention heads ($H$) & 14 \\
Head dim ($d/H$) & 64 \\
Text vocab (Qwen tokenizer) & 151\,936 \\
Codec vocab ($|\mathcal{V}|$) & 6\,561 \\
Max sequence length & 2\,048 \\
Block size ($B$) & 32 \\
\bottomrule
\end{tabular}
\end{table}

\myparagraph{Training Input Format.}
During training, the input sequence consists of $N$ text tokens followed by $L$ codec tokens:
\begin{equation*}
[\texttt{<text>}\, t_1 \cdots t_N \,\texttt{<EOS>}\, c_1 \cdots c_L]
\end{equation*}
Text tokens and codec tokens are embedded via separate embedding layers $\mathbf{E}_{\text{text}} \in \mathbb{R}^{|\mathcal{V}_{\text{text}}| \times d}$ and $\mathbf{E}_{\text{codec}} \in \mathbb{R}^{|\mathcal{V}| \times d}$ into a shared $d$-dimensional space, then concatenated to form the input $\mathbf{h}_0 \in \mathbb{R}^{(N+L) \times d}$. During each training step, codec tokens are randomly masked according to the diffusion timestep $t$ (Eq.~\ref{eq:masking}), and the model learns to reconstruct the original tokens.

\myparagraph{Inference Input Format.}
For zero-shot speaker adaptation, we employ a reference-and-generation paradigm:
\begin{equation*}
\begin{aligned}
[\texttt{<ref\_text>}\, \texttt{<gen\_text>}\, \texttt{<EOS>}\, r_1 &\cdots r_M \, \texttt{[MASK]}^L]
\end{aligned}
\end{equation*}
The reference text is the transcript of a 3--5\,s speaker prompt. The reference codec tokens ($M = 150$--$250$ at $f_r = 50$\,Hz) encode the speaker's voice characteristics. The generation section ($L$ tokens) starts fully masked and is iteratively denoised through block diffusion (Eq.~\ref{eq:blocks}), providing the model with both linguistic content and speaker identity through attention-based conditioning. The maximum sequence length is 2048 tokens ($\sim$40\,s of speech).

\subsection{Inference}
\label{sec:inference}

At inference time, we generate speech through sequential block diffusion decoding. Starting from a fully masked generation segment, we process $K$ blocks sequentially, applying $T$ denoising steps within each block (default $T = B/2 = 16$).

\myparagraph{Confidence-Based Sampling.}
At each denoising step $s \in \{1, \ldots, T\}$ within a block, the model predicts token distributions for all masked positions. For each masked position $i \in \mathcal{M}_s$, we compute:
\begin{equation}
\hat{x}_i = \arg\max_{v \in \mathcal{V}} p_\theta(x_i = v \mid \mathbf{z}_s, \mathbf{c})
\label{eq:decode}
\end{equation}
Position $i$ is unmasked if the model confidence exceeds a threshold $\tau$:
\begin{equation}
\mathcal{U}_s = \big\{i \in \mathcal{M}_s : \max_{v} p_\theta(x_i = v \mid \mathbf{z}_s, \mathbf{c}) > \tau \big\}
\label{eq:unmask}
\end{equation}
with $\tau = 0.6$. Unmasked positions are fixed to $\hat{x}_i$; remaining positions stay masked for subsequent steps. This allows the model to commit to high-certainty tokens first, then resolve ambiguous positions.

\myparagraph{Early Stopping.}
If all masked positions are unmasked ($\mathcal{M}_s = \emptyset$) before step $T$, denoising terminates early, reducing computation for easy blocks.

\myparagraph{Generation Speed.}
Each block of $B = 32$ tokens spans $B / f_r = 0.64$\,s of audio. Blocks are decoded sequentially with up to $T$ parallel denoising steps each, yielding an RTF of 0.15 at $T = 16$. KV caching across blocks reduces latency for subsequent blocks, and the block-sequential design enables streaming with low time-to-first-audio.
\section{Experiments}

\subsection{Experimental Setup}

\myparagraph{Training Data.}
We train DLLM-TTS through a two-stage curriculum:
\textit{Stage~1} trains the model on 16K hours sampled from the Emilia dataset~\cite{he2024emilia} for 20 epochs, establishing coherent codec token generation conditioned on text and speaker prompts. Training uses a batch size of 16 per GPU with 8-step gradient accumulation (effective batch size 128) across $8 \times$H100 GPUs for 3 days.
\textit{Stage~2} fine-tunes on 4K hours of high-quality synthetic speech to improve prosody and alignment.
Throughout both stages, we sample $t \sim \mathcal{U}[0,1]$ and apply per-token masking with probability $1 - \alpha_t$ (Eq.~\ref{eq:masking}). We use AdamW~\cite{loshchilov2019adamw} with learning rate $1 \times 10^{-4}$, cosine schedule with 1\% warmup, and effective batch size 128.

\myparagraph{Evaluation.}
We evaluate on the \textbf{Seed-TTS-eval} benchmark~\cite{anastassiou2024seedtts}, a zero-shot TTS evaluation suite covering standard and challenging scenarios (rare words, complex prosody, long-form utterances).

\myparagraph{Metrics.}
We report:
\begin{itemize}
\item \textbf{Word Error Rate (WER)} and \textbf{Character Error Rate (CER)} computed using Whisper-large-v3~\cite{radford2023whisper} to measure intelligibility.
\item \textbf{Speaker Similarity (SIM)} measured as cosine similarity between speaker embeddings extracted using WavLM-TDNN~\cite{chen2022wavlm}, evaluating voice cloning fidelity.
\item \textbf{Mean Opinion Score (MOS)} collected from 25 listeners following the CodecMOS-Accent protocol~\cite{huang2026codecmos}, assessing perceptual naturalness and prosody on a 5-point scale.
\end{itemize}

\myparagraph{Baselines.}
We compare against autoregressive models (LLASA~\cite{ye2025llasa}, IndexTTS2~\cite{zhou2025indextts2}, Qwen2.5-Omni~\cite{xu2025qwenomni}), non-autoregressive models (F5-TTS~\cite{chen2024f5tts}, MaskGCT~\cite{wang2024maskgct}, CosyVoice3~\cite{du2025cosyvoice3}, OpenAudio-s1-mini~\cite{openaudio2024}), and hybrid models (DiTAR~\cite{jia2025ditar}).

\subsection{Results and Analysis}

Table~\ref{tab:main} presents results on the Seed-TTS-eval benchmark. With $T=32$ denoising steps and block size $B=32$, \textbf{DLLM-TTS achieves a WER of 2.25\% and one of the highest speaker similarity scores (0.750) among open-source systems}, using only \textbf{0.6B parameters} and \textbf{20K hours of training data}, substantially less than systems trained on up to 250K hours. These results demonstrate that block discrete diffusion achieves competitive intelligibility while providing strong zero-shot voice cloning.

\myparagraph{Subjective Quality.}
On subjective evaluation (Table~\ref{tab:main}), DLLM-TTS attains a MOS of \textbf{4.25}, behind only OpenAudio-s1-mini and LLASA-3B and ahead of all other baselines. This confirms that our objective gains translate into perceptual quality, and that block-wise denoising boundaries do not harm naturalness or prosody.

\myparagraph{Data Efficiency.}
Compared to autoregressive codec language models trained on 60K--250K hours, DLLM-TTS achieves competitive intelligibility with only \textbf{20K hours}, a \textbf{3--12$\times$ data reduction}. We attribute this to the masked diffusion training objective (Eq.~\ref{eq:loss}), which exposes each sequence to diverse masking patterns across timesteps, providing implicit data augmentation compared to the single left-to-right ordering of autoregressive training.

\myparagraph{Latency.}
Following the inference procedure in Section~\ref{sec:inference}, with $T=16$ steps per block and confidence threshold $\tau=0.6$, DLLM-TTS achieves an RTF of \textbf{0.15}. Each block of $B=32$ codec tokens spans $B/f_r = 0.64$\,s of audio, enabling low time-to-first-audio and streaming synthesis after the first block is denoised. KV caching across the $K$ sequential blocks further reduces latency.

\begin{table}[t]
\centering
\caption{Results on Seed-TTS-eval (English). MOS collected from 25 listeners under the CodecMOS-Accent protocol. \textbf{Bold}: best. \underline{Underline}: second best.}
\label{tab:main}
\renewcommand{\arraystretch}{1.0}
\resizebox{\columnwidth}{!}{
\begin{tabular}{lccccc}
\toprule
Model & Params & WER$\downarrow$ & CER$\downarrow$ & SIM$\uparrow$ & MOS$\uparrow$ \\
\midrule

\multicolumn{6}{l}{\textit{Autoregressive Models}} \\

LLASA-3B & 3B & 3.14 & 1.59 & 0.579 & \underline{4.28} \\
Qwen2.5-Omni & 7B & 2.72 & 1.70 & 0.632 & 3.85 \\

\midrule
\multicolumn{6}{l}{\textit{Non-Autoregressive Models}} \\

F5-TTS & 0.3B & 2.00 & 1.53 & 0.670 & 4.05 \\
MaskGCT & -- & 2.62 & 2.27 & 0.717 & 3.76 \\
OpenAudio-s1-mini & 0.5B & \underline{1.94} & 1.18 & 0.550 & \textbf{4.29} \\

\midrule
\multicolumn{6}{l}{\textit{Hybrid Models}} \\

DiTAR & 0.6B & \textbf{1.69} & \textbf{1.02} & \underline{0.735} & 4.20 \\
CosyVoice3 & 0.5B & 2.02 & 1.16 & 0.718 & 4.10 \\
IndexTTS2 & 1.5B & 2.23 & 1.08 & 0.706 & 3.98 \\

\midrule
\multicolumn{6}{l}{\textit{Block Diffusion (Ours)}} \\

\textbf{DLLM-TTS} & \textbf{0.6B} & 2.25 & \underline{1.05} & \textbf{0.750} & 4.25 \\

\bottomrule
\end{tabular}
}
\end{table}

\begin{figure}
    \centering
    \includegraphics[width=0.82\linewidth]{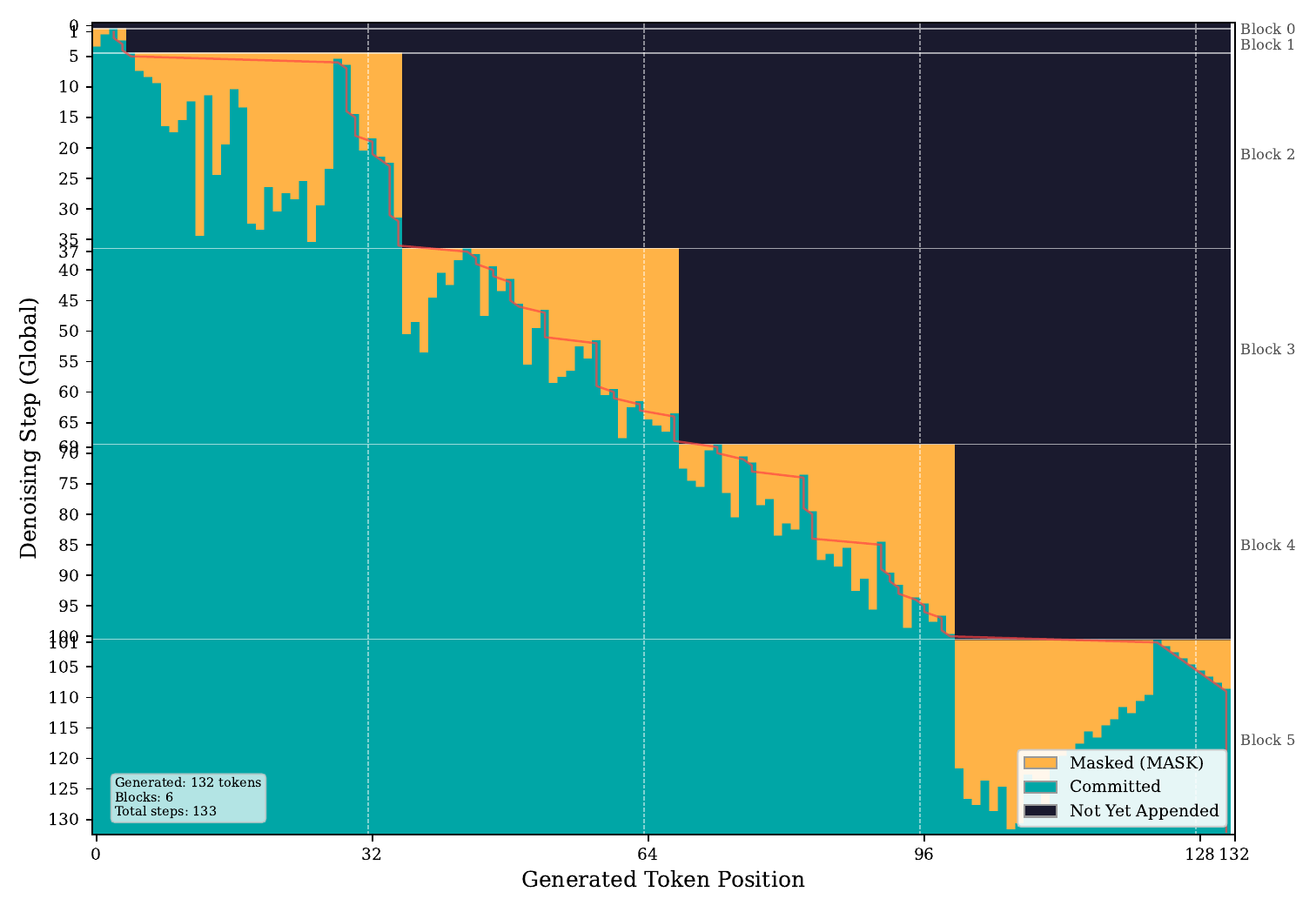}
    \caption{Visualization of the block-wise denoising process. ($B = 32$, $T = 32$)}
    \label{fig:denoising}
\end{figure}

\subsection{Ablation Studies}

\myparagraph{Effect of Denoising Steps.}
Table~\ref{tab:ablation} (top) evaluates the number of denoising steps $T$ per block during inference (Section~\ref{sec:inference}). With $\tau=0.6$ (Eq.~\ref{eq:unmask}), increasing $T$ improves intelligibility while maintaining stable speaker similarity: WER decreases from 14.58\% to 2.25\% and CER from 7.86\% to 1.05\% when moving from $T=8$ to $T=32$, while SIM remains relatively stable (0.746--0.765). At $T=16$ ($=B/2$), the model achieves a favorable speed--quality trade-off with an RTF of \textbf{0.15}.

\begin{table}[t]
\centering
\caption{Ablation studies. \textit{Top}: denoising steps $T$ (at $B=32$). \textit{Bottom}: block size $B$ (at $T=B$). \textbf{Bold}: best.}
\label{tab:ablation}
\renewcommand{\arraystretch}{1.0}
\begin{tabular}{cccc}
\toprule
Config & WER$\downarrow$ & CER$\downarrow$ & SIM$\uparrow$ \\
\midrule
\multicolumn{4}{l}{\textit{Denoising steps ($T$), $B=32$}} \\
$T{=}8$  & 14.58 & 7.86 & 0.748 \\
$T{=}16$ & 3.84  & 1.25 & \textbf{0.765} \\
$T{=}32$ & \textbf{2.25} & \textbf{1.05} & 0.750 \\
$T{=}64$ & 8.83  & 6.76 & 0.746 \\
\midrule
\multicolumn{4}{l}{\textit{Block size ($B$), $T=B$}} \\
$B{=}8$  & 4.19 & 2.06 & 0.725 \\
$B{=}16$ & 3.04 & 1.43 & 0.746 \\
$B{=}32$ & \textbf{2.25} & \textbf{1.05} & \textbf{0.750} \\
\bottomrule
\end{tabular}
\end{table}

\myparagraph{Effect of Block Size.}
Table~\ref{tab:ablation} (bottom) varies block size $B$ with matched denoising steps ($T=B$). Larger $B$ raises within-block parallelism but reduces sequential boundaries $K=\lceil L/B\rceil$, weakening cross-block conditioning under the staircase mask (Eq.~\ref{eq:attn}); smaller $B$ strengthens autoregressive guidance but limits parallel context. Empirically, $B=32$ is best (WER 2.25\%, CER 1.05\%, SIM 0.750): smaller blocks ($B=8,16$) raise error rates, while excessively large blocks sacrifice sequential modeling capacity and degrade intelligibility. Speaker similarity stays stable across configurations (0.725--0.750), showing robust voice cloning.
\section{Conclusion}

We presented DLLM-TTS, a TTS framework based on block discrete diffusion language modeling over neural audio codec tokens. By decomposing codec token sequences into blocks and applying masked diffusion within each block while processing them sequentially, the model learns both local acoustic coherence and global text–speech alignment without requiring explicit duration modeling or phoneme-level annotations.

A 0.6B-parameter model trained on 20K hours of data achieves competitive intelligibility on the Seed-TTS benchmark while obtaining strong speaker similarity, demonstrating improved data efficiency compared to autoregressive codec language models trained on 60K--250K hours. The block-parallel inference strategy achieves a real-time factor (RTF) of 0.15, enabling practical real-time speech synthesis.

These results suggest that block discrete diffusion language models offer a scalable alternative to autoregressive speech models, combining the parallel generation advantages of diffusion with the sequential structure needed for stable text–speech alignment.
\clearpage

\section{Use of Generative AI Disclosure}
In preparing this manuscript, the authors used generative AI tools for language refinement (rephrasing and improving the clarity of author-written text) and as a coding assistant (helping write and debug software for experiments and analysis). All research contributions, including the methodology, experimental design, results, and scientific claims, are the authors' own. The authors reviewed and verified all AI-assisted text and code, and take full responsibility for the content of this paper.


\bibliographystyle{IEEEtran}
\bibliography{mybib}

@article{defossez2022encodec,
  title={High Fidelity Neural Audio Compression},
  author={D{\'e}fossez, Alexandre and Copet, Jade and Synnaeve, Gabriel and Adi, Yossi},
  journal={arXiv preprint arXiv:2210.13438},
  year={2022}
}

@article{ye2024xcodec,
  title={Codec Does Matter: Exploring the Semantic Shortcoming of Codec for Audio Language Model},
  author={Ye, Zhen and Ai, Peiwen and Sun, Jiahe and others},
  journal={arXiv preprint arXiv:2408.17175},
  year={2024}
}

@article{wang2023valle,
  title={Neural Codec Language Models are Zero-Shot Text to Speech Synthesizers},
  author={Wang, Chengyi and Chen, Sanyuan and Wu, Yu and Zhang, Ziqiang and Zhou, Long and Liu, Shujie and Chen, Zhuo and Liu, Yanqing and Wang, Huaming and Li, Jinyu and others},
  journal={arXiv preprint arXiv:2301.02111},
  year={2023}
}

@article{chen2024valle2,
  title={{VALL-E} 2: Neural Codec Language Models are Human Parity Zero-Shot Text to Speech Synthesizers},
  author={Chen, Sanyuan and Yu, Shujie and Zhou, Long and Wu, Yu and others},
  journal={arXiv preprint arXiv:2406.05370},
  year={2024}
}

@article{ye2025llasa,
  title={{LLASA}: Scaling Train-Time and Inference-Time Compute for {LLaMA}-based Speech Synthesis},
  author={Ye, Zhen and Ai, Peiwen and Sun, Jiahe and others},
  journal={arXiv preprint arXiv:2502.04128},
  year={2025}
}

@inproceedings{ju2024naturalspeech3,
  title={{NaturalSpeech} 3: Zero-Shot Speech Synthesis with Factorized Codec and Diffusion Models},
  author={Ju, Zeqian and Wang, Yuancheng and Shen, Kai and Tan, Xu and Xin, Detai and Yang, Dongchao and Liu, Yanqing and Leng, Yichong and Song, Kaitao and Tang, Siliang and others},
  booktitle={Proc. ICML},
  year={2024}
}

@inproceedings{li2023styletts2,
  title={{StyleTTS} 2: Towards Human-Level Text-to-Speech through Style Diffusion and Adversarial Training with Large Speech Language Models},
  author={Li, Yinghao Aaron and Han, Cong and Raber, Vinay S and Mesgarani, Nima},
  booktitle={Proc. NeurIPS},
  year={2023}
}

@article{chen2024f5tts,
  title={{F5-TTS}: A Fairytaler that Fakes Fluent and Faithful Speech with Flow Matching},
  author={Chen, Yushen and Wu, Zhikang and Zhang, Ziyang and others},
  journal={arXiv preprint arXiv:2410.06885},
  year={2024}
}

@inproceedings{le2024voicebox,
  title={Voicebox: Text-Guided Multilingual Universal Speech Generation at Scale},
  author={Le, Matthew and Vyas, Apoorv and Shi, Bowen and Karrer, Brian and Sager, Leda and Adel, Xintong and Williamson, Michal and Manohar, Vimal and Moritz, Niko and Hsu, Wei-Ning and others},
  booktitle={Proc. NeurIPS},
  year={2023}
}

@article{du2024cosyvoice,
  title={{CosyVoice}: A Scalable Multilingual Zero-shot Text-to-speech Synthesizer Based on Supervised Semantic Tokens},
  author={Du, Zhihao and Chen, Qian and Shi, Shiliang and others},
  journal={arXiv preprint arXiv:2407.05407},
  year={2024}
}

@article{borsos2023soundstorm,
  title={{SoundStorm}: Efficient Parallel Audio Generation},
  author={Borsos, Zal{\'a}n and Sharifi, Matt and Vincent, Damien and Kharitonov, Eugene and Zeghidour, Neil and Tagliasacchi, Marco},
  journal={arXiv preprint arXiv:2305.09636},
  year={2023}
}

@article{jia2025ditar,
  title={{DiTAR}: Diffusion Transformer Autoregressive Modeling for Speech Generation},
  author={Jia, Dongya and Chen, Zhuo and Wang, Yuxuan and others},
  journal={arXiv preprint arXiv:2502.03930},
  year={2025}
}

@article{jiang2025megatts3,
  title={{MegaTTS} 3: Sparse Alignment Enhanced Latent Diffusion Transformer for Zero-Shot Speech Synthesis},
  author={Jiang, Ziyue and Ren, Yi and Li, Ruibin and others},
  journal={arXiv preprint arXiv:2502.18924},
  year={2025}
}

@inproceedings{sahoo2024mdlm,
  title={Simple and Effective Masked Diffusion Language Models},
  author={Sahoo, Subham Sekhar and Arriola, Marianne and Schiff, Yair and Gokaslan, Aaron and Marroquin, Edgar and Chiu, Justin T and Rush, Alexander and Kuleshov, Volodymyr},
  booktitle={Proc. NeurIPS},
  year={2024}
}

@article{nie2025llada,
  title={Large Language Diffusion Models},
  author={Nie, Shen and Zhu, Fengqi and You, Chao and Zhang, Xiaojie and Gong, Jianfeng},
  journal={arXiv preprint arXiv:2502.09992},
  year={2025}
}

@inproceedings{arriola2025bd3lm,
  title={Block Diffusion: Interpolating Between Autoregressive and Diffusion Language Models},
  author={Arriola, Marianne and Gokaslan, Aaron and Sahoo, Subham Sekhar and Hsu, Lili and Kuleshov, Volodymyr},
  booktitle={Proc. ICLR},
  year={2025}
}

@article{he2024emilia,
  title={Emilia: An Extensive, Multilingual, and Diverse Speech Dataset for Large-Scale Speech Generation},
  author={He, Haorui and Shang, Zengqiang and Wang, Chaoren and others},
  journal={arXiv preprint arXiv:2407.05361},
  year={2024}
}

@article{anastassiou2024seedtts,
  title={Seed-{TTS}: A Family of High-Quality Versatile Speech Generation Models},
  author={Anastassiou, Philip and Cheng, Jiaqi and Leng, Dongyi and others},
  journal={arXiv preprint arXiv:2406.02430},
  year={2024}
}

@inproceedings{radford2023whisper,
  title={Robust Speech Recognition via Large-Scale Weak Supervision},
  author={Radford, Alec and Kim, Jong Wook and Xu, Tao and Brockman, Greg and McLeavey, Christine and Sutskever, Ilya},
  booktitle={Proc. ICML},
  year={2023}
}

@article{chen2022wavlm,
  title={{WavLM}: Large-Scale Self-Supervised Pre-Training for Full Stack Speech Processing},
  author={Chen, Sanyuan and Wang, Chengyi and Chen, Zhengyang and Wu, Yu and Liu, Shujie and Chen, Zhuo and Li, Jinyu and Kanda, Naoyuki and Yoshioka, Takuya and Xiao, Xiong and others},
  journal={IEEE Journal of Selected Topics in Signal Processing},
  volume={16},
  number={6},
  pages={1505--1518},
  year={2022}
}

@article{yang2024qwen2,
  title={Qwen2 Technical Report},
  author={Yang, An and Yang, Baosong and Hui, Binyuan and others},
  journal={arXiv preprint arXiv:2407.10671},
  year={2024}
}

@article{su2024rope,
  title={{RoFormer}: Enhanced Transformer with Rotary Position Embedding},
  author={Su, Jianlin and Ahmed, Murtadha and Lu, Yu and Pan, Shengfeng and Bo, Wen and Liu, Yunfeng},
  journal={Neurocomputing},
  volume={568},
  pages={127063},
  year={2024}
}

@inproceedings{loshchilov2019adamw,
  title={Decoupled Weight Decay Regularization},
  author={Loshchilov, Ilya and Hutter, Frank},
  booktitle={Proc. ICLR},
  year={2019}
}

@article{du2025cosyvoice3,
  title={{CosyVoice} 3: Towards In-the-Wild Speech Generation via Scaling-Up and Post-Training},
  author={Du, Zhihao and Gao, Changfeng and Wang, Yuxuan and others},
  journal={arXiv preprint arXiv:2505.17589},
  year={2025}
}

@article{openaudio2024,
  title={{Fish-Speech}: Leveraging Large Language Models for Advanced Multilingual Text-to-Speech Synthesis},
  author={Liao, Shijia and Wang, Yuxuan and Li, Tianyu and Cheng, Yifan and Zhang, Ruoyi and Zhou, Rongzhi and Xing, Yijin},
  journal={arXiv preprint arXiv:2411.01156},
  year={2024}
}

@article{wang2024maskgct,
  title={{MaskGCT}: Zero-Shot Text-to-Speech with Masked Generative Codec Transformer},
  author={Wang, Yuancheng and Zhan, Haoyue and Liu, Liwei and Zeng, Ruihong and Guo, Haotian and Zheng, Jiachen and Zhang, Qiang and Zhang, Xueyao and Zhang, Shunsi and Wu, Zhizheng},
  journal={arXiv preprint arXiv:2409.00750},
  year={2024}
}

@article{zhou2025indextts2,
  title={{IndexTTS2}: A Breakthrough in Emotionally Expressive and Duration-Controlled Auto-Regressive Zero-Shot Text-to-Speech},
  author={Zhou, Siyi and Zhou, Yiquan and He, Yi and Zhou, Xun and Wang, Jinchao and Deng, Wei and Shu, Jingchen},
  journal={arXiv preprint arXiv:2506.21619},
  year={2025}
}

@article{xu2025qwenomni,
  title={{Qwen2.5-Omni} Technical Report},
  author={Xu, Jin and Guo, Zhifang and He, Jinzheng and others},
  journal={arXiv preprint arXiv:2503.20215},
  year={2025}
}

@article{huang2026codecmos,
  title={{CodecMOS-Accent}: A {MOS} Benchmark of Resynthesized and {TTS} Speech from Neural Codecs Across {English} Accents},
  author={Huang, Wen-Chin and Sanders, Nathan and Cooper, Erica},
  journal={arXiv preprint arXiv:2603.14328},
  year={2026}
}

\end{document}